\documentclass[10pt,twocolumn,letterpaper]{article}

\usepackage{iccv}
\usepackage{times}
\usepackage{epsfig}
\usepackage{graphicx}
\usepackage{amsmath}
\usepackage{amssymb}
\usepackage{subfigure}
\usepackage{bm}

\usepackage[pagebackref=true,breaklinks=true,letterpaper=true,colorlinks,bookmarks=false]{hyperref}

\iccvfinalcopy 

\def\iccvPaperID{10} 

\ificcvfinal\fi
\begin{document}
\title{Explaining Convolutional Neural Networks \\ using Softmax Gradient Layer-wise Relevance Propagation}

\author{Brian Kenji Iwana\thanks{Equal contribution} \quad\quad Ryohei Kuroki\footnotemark[1] \quad\quad Seiichi Uchida\\
Kyushu University, Fukuoka, Japan\\
{\tt\small brian@human.ait.kyushu-u.ac.jp, ryohei.kuroki@gmail.com, uchida@ait.kyushu-u.ac.jp}
}

\maketitle
\thispagestyle{empty}

\begin{abstract}
Convolutional Neural Networks (CNN) have become state-of-the-art in the field of image classification. However, not everything is understood about their inner representations. This paper tackles the interpretability and explainability of the predictions of CNNs for multi-class classification problems. Specifically, we propose a novel visualization method of pixel-wise input attribution called Softmax-Gradient Layer-wise Relevance Propagation (SGLRP). The proposed model is a class discriminate extension to Deep Taylor Decomposition (DTD) using the gradient of softmax to back propagate the relevance of the output probability to the input image. Through qualitative and quantitative analysis, we demonstrate that SGLRP can successfully localize and attribute the regions on input images which contribute to a target object's classification. We show that the proposed method excels at discriminating the target objects class from the other possible objects in the images. We confirm that SGLRP performs better than existing Layer-wise Relevance Propagation (LRP) based methods and can help in the understanding of the decision process of CNNs. 

\end{abstract}

\section{Introduction}

Artificial Neural Networks (ANN) have become a staple in machine learning and pattern recognition~\cite{schmidhuber2015deep} due to their success in image~\cite{he2015delving,wan2013regularization} and text~\cite{uchida2016further} classification. 
However, despite their success, Convolutional Neural Networks (CNN)~\cite{lecun1998gradient} have been historically regarded as ``black boxes"~\cite{Zeiler_2014,yosinski2015understanding,grun2016taxonomy}.
In particular, deep CNNs, such as the Visual Geometry Group (VGG)~\cite{Simonyan14c} networks, have millions of parameters making it difficult to understand the internal representations and decision processes of the networks.

\begin{figure}[t]
        \centering
    \subfigure[Forward]{
        \centering
        \includegraphics[width=0.9\linewidth]{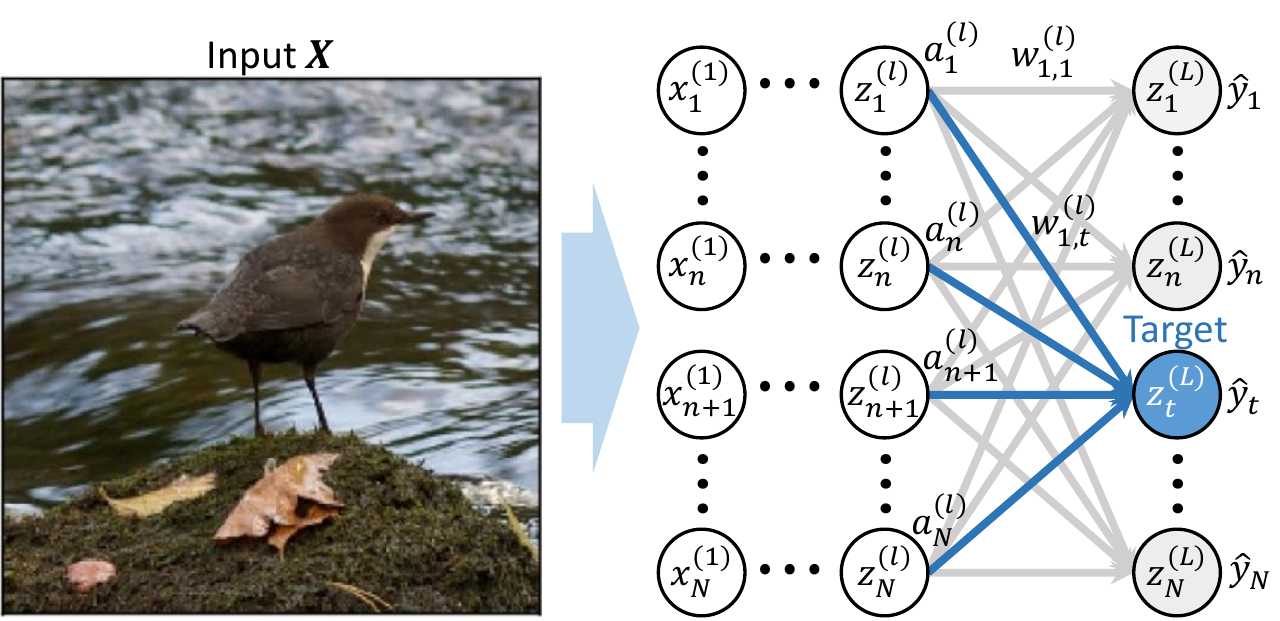}
    }
    \subfigure[The Proposed SGLRP]{
        \centering
        \includegraphics[width=0.9\linewidth]{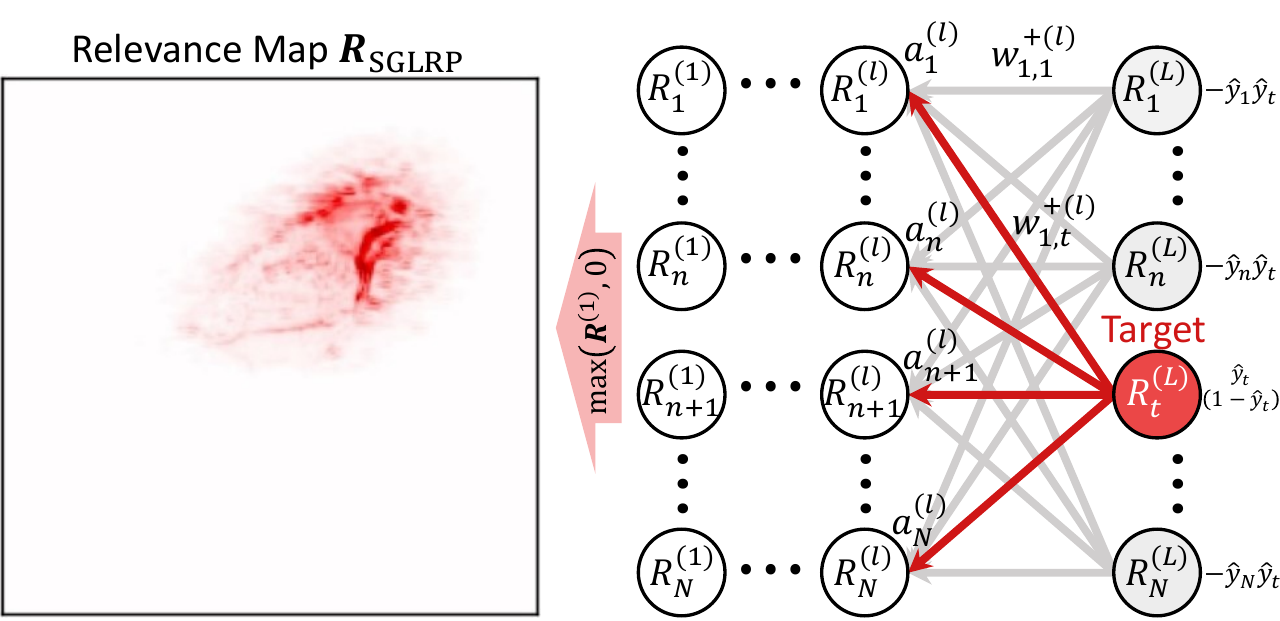}
        }
        \caption{Our SGLRP targets a specific output node $t$ corresponding to a class and determines the pixel-wise relevance using the gradient of softmax. (a) is the forward pass of a neural network where $\bm{X}=x_1^{(1)},\dots,x_n^{(1)},\dots,x_N^{(1)}$ is the input, $z_n^{(l)}$ is the intermediate value of node $n$ in layer $l$, $a_n^{(1)}$ is the activation, and $w_{1,t}^{(l)}$ is the weight between $a_1^{(l)}$ and $z_t^{(L)}$. $\hat{y}_n$ is the result from softmax for node $n$ and $\hat{y}_t$ is the result for target node $t$. (b) is the relevance propagation of the gradient of softmax through $R_n^{(l)}$ to the relevance map $\bm{R}_{\mathrm{SGLRP}}$. }\label{sglrp_target_figure}
\end{figure}

Thus, there have been many recent attempts to try and visualize, explain, and understand the inner workings of CNNs~\cite{grun2016taxonomy}. 
For example, there have been attempts to understand CNNs by manipulating the input images through occlusion~\cite{Zeiler_2014} and optimized activation excitement~\cite{olah2017feature}. 
There have also been many works that aim to elucidate the internal decision processes by tracing the operations of a CNN to produce input contribution heatmaps. 
Some examples include, Deconvolutional Networks (DeconvNets)~\cite{Simonyan2013DeepIC,Zeiler_2014}, Layer-wise Relevance Propagation (LRP)~\cite{Bach2015,Montavon2017a}, Guided Back Propagation~\cite{Springenberg2015}, and Class Activation Mapping (CAM)~\cite{Lin2014, Zhou_2016_CVPR,Selvaraju2017}.
Beyond understanding CNNs, contribution maps can also be used for weakly supervised object localization~\cite{Bazzani_2016,Teh_2016,Tang_2018}.

In this paper, we propose a novel method to visualize the input's pixel-wise contribution toward the object's classification. 
Specifically, we propose an extension of LRP called Softmax-Gradient Layer-wise Relevance Propagation (SGLRP). 
As shown in Fig.~\ref{sglrp_target_figure}, the proposed method does this by propagating the gradient of softmax through a trained network in order to realize robust class discriminant relevance maps. 

SGLRP differs from LRP in that we propose the use of the gradient of softmax as the output layer's relevance (the starting propagated signal). 
By using the gradient of softmax, SGLRP subtracts the relevance of non-target classes based on the probability that the image is classified as that class. 
This allows SGLRP to specifically remove surrounding, conflicting, or occluding objects not related to the target class. 
Due to this, the proposed method is an effective method of determining the pixel-wise contribution that the input image has on its classification.

The contributions of this paper are as follows.
\begin{itemize}
    \item We propose a novel and efficient way to visualize the pixel-wise contributions toward the specific classifications of objects by using the gradient of softmax in LRP. 
    \item Through qualitative evaluations, we show that the proposed method can produce class discriminative relevance maps. We show that SGLRP can localize contributing regions and objects and is robust even when multiple different classes are present. 
    \item We demonstrate through quantitative evaluations that the proposed method performs better than other LRP-based visualizations. To evaluate the proposed method, we use an object pointing game and a patch removal evaluation.
\end{itemize}

\section{Related Work}\label{related_works}

Neural network visualization techniques can be broadly separated into three categories, input modification methods, back propagation-based methods, and CAM methods. 

\subsection{Input Modification Methods}

Input modification methods observe the changes in the outputs based on changes to the inputs. 
This can be done using perturbed variants~\cite{Alipanahi_2015,Zhou_2015}, masks~\cite{Zeiler_2014}, or noise~\cite{zhou2014object}. 
For instance, Zeiler and Fergus~\cite{Zeiler_2014} create heatmaps based on the drop in prediction probability from input images with masked patches at different places. 
The problem with these methods is that they require exhaustive input modifications and can be computationally costly. 

Another method of exploring the internal representations of CNNs is to excite particular nodes in order to derive class caricaturizations. 
For example, Mahendran and Vedaldi~\cite{Mahendran_2016} modify pre-images and Olah et al.~\cite{olah2017feature} optimize generated images to maximally activate particular neurons. 

\subsection{Back Propagation-based Methods}

Back propagation-based methods trace the contribution of the output backwards through the network to the input. 
In the classic example, DeconvNet~\cite{Zeiler_2014,Simonyan2013DeepIC} back propagates the output through the network. 
However, instead of using the Rectified Linear Unit (ReLU) from the forward pass, DeconvNet applies ReLU to the back propagated signal. 
Guided Back Propagation~\cite{Springenberg2015} is similar to DeconvNet except it blocks the negative values from for the forward and the backwards signals.


Deep Taylor Decomposition (DTD)~\cite{Montavon2017a} based models such as LRP~\cite{Bach2015} and Contrastive LRP (CLRP)~\cite{gu2018understanding} use both the gradient and the input to propagate the relevance of the output backwards through the trained network. 
In this way, the LRP models redistribute the output backwards through the network in order to determine the contribution that nodes have on the classification. 
PatternNet and PatternAttribution~\cite{kindermans2018learning} uses a trained back propagation method which yields a similar target to LRP. 
LRP and DTD has been used for applications such as video~\cite{Gan_2015}, book classification explanation~\cite{Jolly_2018}, and bioinformatics~\cite{Sturm_2016}. 

Using the gradients of activation functions to back propagate relevance can sometimes lead to misleading contribution attributions due to discontinuous or vanishing gradients. 
In order to overcome this, instead of using the gradients to back propagate the relevance, Deep Learning Important FeaTures (DeepLIFT)~\cite{shrikumar2017learning} uses the difference between the activations of reference inputs. 
SHapeley Additive exPlanation (SHAP)~\cite{NIPS2017_7062} extends DeepLIFT to include Shapely approximations. 
In addition, another way to solve this problem is the use of Integrated Gradients~\cite{sundararajan2017axiomatic,qi2019visualizing}.


\subsection{Class Activation Mapping Methods}

Class Activation Mapping (CAM)~\cite{Lin2014, Zhou_2016_CVPR} methods combine the output and the weights between a Global Average Pooling (GAP) layer and the output layer to combine the feature maps of the last convolutional layer into a heatmap. 
The downside of CAM is that it requires a final GAP and cannot be applied to every network. 

Gradient-Weighted CAM (Grad-CAM)~\cite{Selvaraju2017} is a generalization of CAM that can target any layer and introduces the gradient information to CAM. 
The problem with CAM-based methods is that they specifically target high-level layers. 
Therefore, hybrid methods, such as Guided Grad-CAM~\cite{Selvaraju2017}, combine the qualities of CAMs with pixel-wise methods, such as guided back propagation. 
 
\subsection{Miscellaneous Methods}

There are also many other visualization and explanation methods for neural networks. 
For example, by observing the maximal activations of layers, it is possible to visualize contributing regions of feature maps~\cite{Zeiler_2014,yosinski2015understanding} or the use of attention visualization~\cite{fukui2019attention,xie2019deep}. 
In addition, the hidden layers of networks can be analyzed using lower-dimensional representations, such as dimensionality reduction~\cite{Rauber_2017}, Relative Neighbor Graphs (RNG)~\cite{Ide_2017}, matrix factorization~\cite{olah2018the}, and modular representation by community detection~\cite{Watanabe_2018}.

\section{Relevance Propagation}
\subsection{Layer-wise Relevance Propagation}\label{lrp}


\begin{figure}[t]
        \centering
        \includegraphics[width=0.9\linewidth]{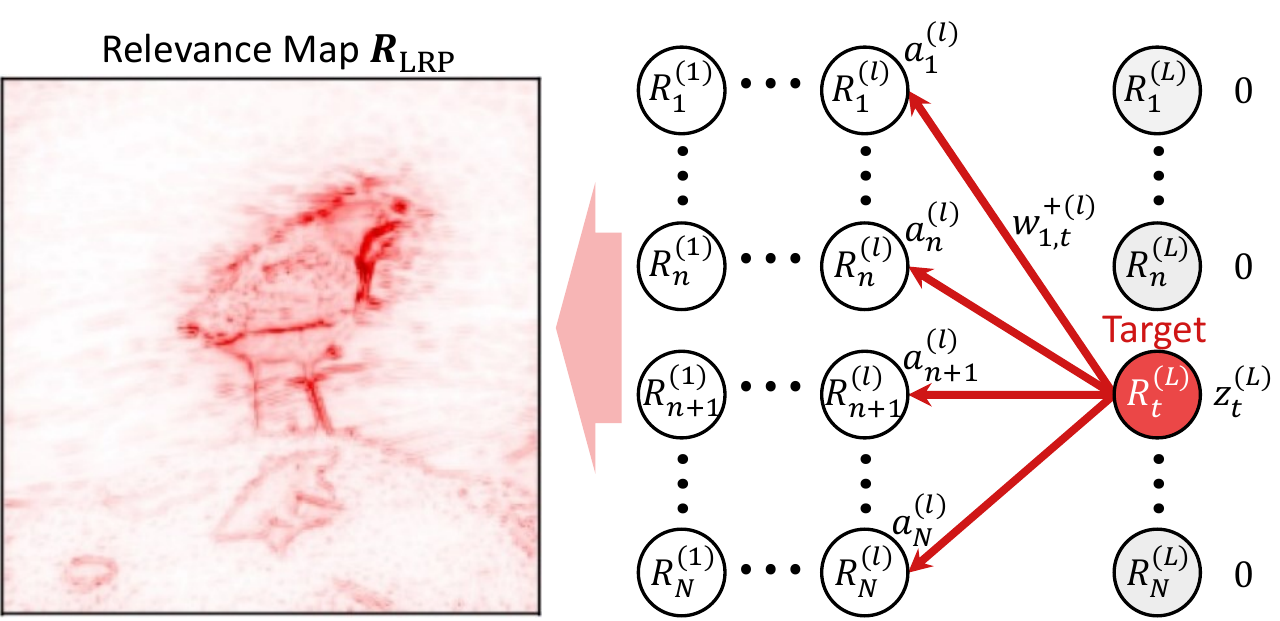}
        \caption{Illustration of LRP}\label{lrp_target_figure}
\end{figure}

LRP~\cite{Bach2015} is based on the idea that the likelihood of a class can be traced backwards through a network to the individual layer-wise nodes or elements of the input.  
Specifically, the contribution, or relevance, to the target output node is back propagated toward the input image creating a map of which pixels contributed to the node. 

An image $\bm{X}$ is classified in a forward pass (Fig.~\ref{sglrp_target_figure}~(a)) and then the relevance $R_{t}^{(L)}$ is propagated backwards through the network using DTD in order to generate a relevance map $\bm{R}_{\mathrm{LRP}}$ (Fig.~\ref{lrp_target_figure}). 
For each layer $l$ in a network with $L$ total layers, $1,\dots,n,\dots,N$ are the nodes in layer $l$ and $1,\dots,m,\dots,M$ are the nodes in layer $l+1$, the relevance $R_{n}^{(l)}$ at node $n$ in layer $l$ is recursively defined by:
\begin{equation}
  R_{n}^{(l)}=
  \sum_m \frac{a_n^{(l)}w_{n,m}^{+(l)}}
  {\sum_{n'}a_{n'}^{(l)}w_{n',m}^{+(l)}}R_m^{(l+1)},
  \label{eq:define_rij_dtd_zplus}
\end{equation}
for nodes using positive semi-definite activation functions, such as ReLU, and: 
\begin{equation}
  R_{n}^{(l)}=
  \sum_m \frac{x_n^{(l)}w_{n,m}^{(l)}-b_n^{(l)}w_{n,m}^{+(l)}-h_n^{(l)}w_{n,m}^{-(l)}}
  {\sum_{n'}x_{n'}^{(l)}w_{n',m}^{(l)}-b_{n'}^{(l)}w_{n',m}^{+(l)}-h_{n'}^{(l)}w_{n',m}^{-(l)}}
  R_m^{(l+1)},
  \label{eq:define_rij_dtd_zbeta}
\end{equation}
for cases where there can be negative values, such as the input, and
\begin{equation}
  R_n^{(L)} = 
  \begin{cases}
    z_t^{(L)} & n=t,\\
    0 & \mathrm{otherwise}, 
  \end{cases}
  \label{relevance_general2}\\
\end{equation}
for the output layer. 
$a_n^{(l)}$ is the post-activation output of node $n$ in layer $l$ and $z_t^{(L)}$ is the pre-softmax value of target node $t$. The range $[b_n^{(l)},h_n^{(l)}]$ represents the lower and the upper limits of $z_n^{(l)}$, respectively. 
Finally, $w_{n,m}^{+(l)}$ and $w_{n,m}^{-(l)}$ are:
\begin{equation}
  w_{n,m}^{+(l)}= \max(w_{n,m}^{(l)},0), 
  \label{eq:weightpositive}
\end{equation}
and
\begin{equation}
  w_{n,m}^{-(l)}= \min(w_{n,m}^{(l)},0),
  \label{eq:weightnegative}
\end{equation}
respectively. 
Using Eq.~\eqref{eq:define_rij_dtd_zbeta} at the input layer producing a pixel-wise relevance map $\bm{R}_{\mathrm{LRP}}$.

\begin{figure}[t]
        \centering
    \subfigure[Forward]{
        \centering
        \begin{minipage}{0.47\columnwidth}
        \centering
            \includegraphics[width=.9\columnwidth]{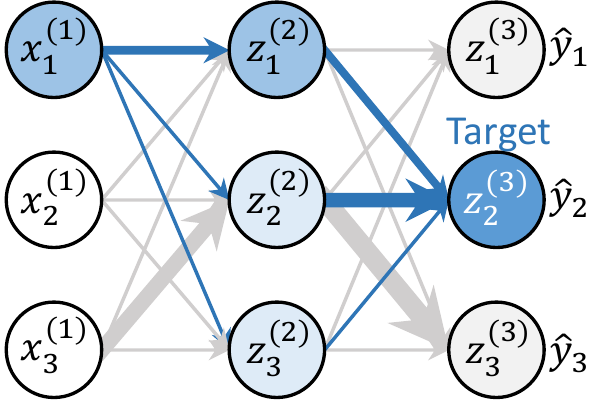}
    \vspace{1mm}
        \end{minipage}
    }
    \subfigure[Relevance Propagation]{
        \centering
        \begin{minipage}{0.47\columnwidth}
        \centering
            \includegraphics[width=.9\columnwidth]{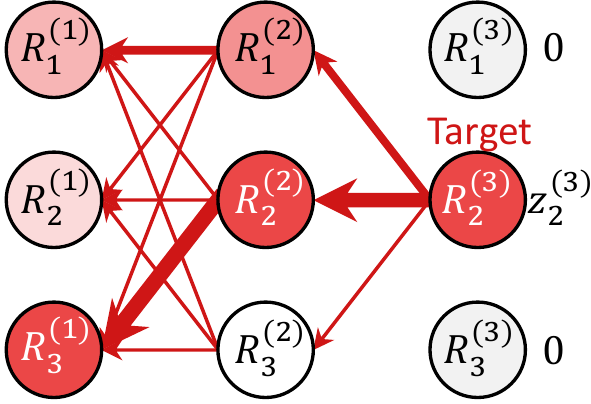}
    \vspace{1mm}
        \end{minipage}
    }
    \caption{In this example, $z_2^{(2)}$ contributes to both $z_2^{(3)}$ and $z_2^{(3)}$ meaning that $z_2^{(2)}$ is not specifically relevant to either. However, LRP deems $z_2^{(2)}$ the most relevant irrespective to the contribution to the non-target node. The thickness of the arrows indicate the contribution to the output in (a) and the relevance propagated in (b). }
    \label{flirt}
\end{figure}

While LRP has been used successfully on interpreting CNNs in various applications~\cite{Bach2015,Sturm_2016,Jolly_2018}, LRP only takes the target class into consideration for the calculation which can lead to miss-attribution of input regions to the relevance~\cite{gu2018understanding}.
For instance, consider the toy network shown in Fig.~\ref{flirt}. 
In the example, input $z_2^{(2)}$ has a large contribution to {\em both} outputs $\hat{y}_2$ and $\hat{y}_3$; this means that $z_2^{(2)}$ is not specifically relevant to neither class 1 nor class 2. 
The actual discriminating factor between the classes is the value from $z_1^{(2)}$. 
However, due to the formulation of LRP, the node at $z_2^{(2)}$ and $x_3^{(1)}$ are determined to be the most relevant when in reality, $x_1^{(1)}$ lead to the value of $\hat{y}_2$. 
In this case, LRP will not be able to determine the actual importance of the inputs.

\subsection{Contrastive Layer-wise Relevance Propagation}\label{clrp}

\begin{figure}[t]
        \centering
        \includegraphics[width=0.9\linewidth]{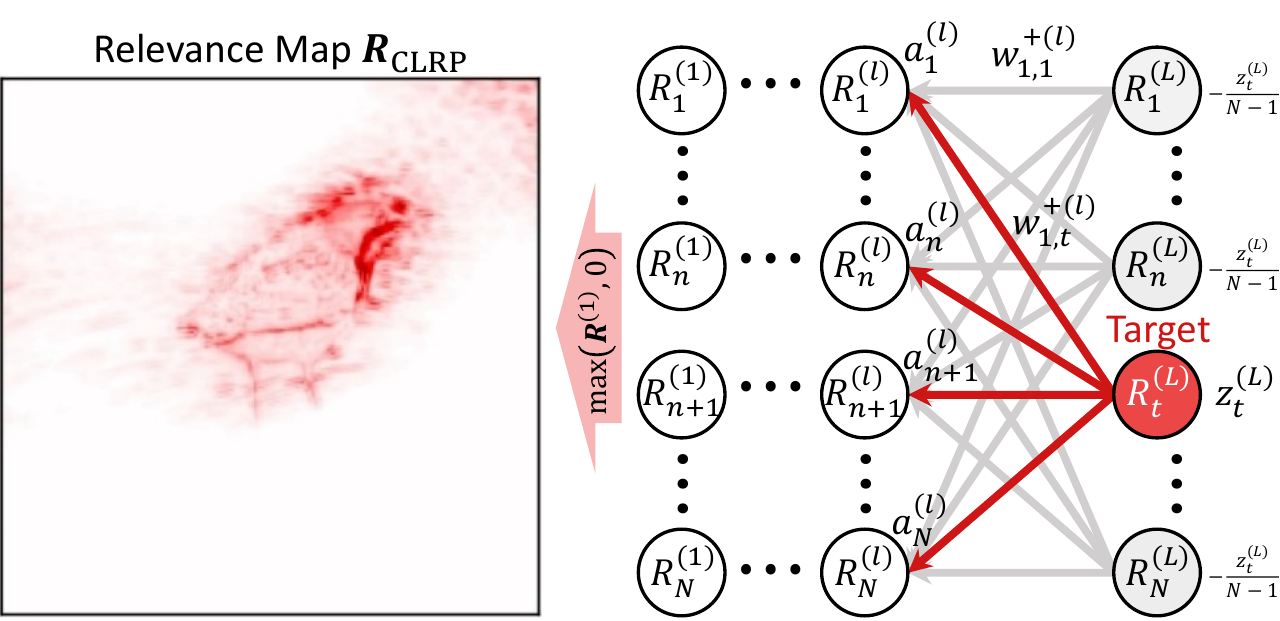}
        \caption{Illustration of CLRP}\label{clrp_target_figure}
\end{figure}

CLRP~\cite{gu2018understanding}, a class contrastive improvement on LRP, was proposed to tackle LRP's insufficient ability for discriminating the target object's class with the non-target classes. 
To do this, as shown in Fig.~\ref{clrp_target_figure}, CLRP subtracts the relevance for non-target classes from the relevance propagation. 
Specifically, the relevance map $\bm{R}_{\mathrm{CLRP}}$ is:
\begin{align}
\begin{split}
    \bm{R}_{\mathrm{CLRP}} =& \max\left(R_1^{(1)},\dots,R_n^{(1)},\dots, R_N^{(1)}, 0\right), 
    \end{split}
    \label{clrp_def}
\end{align}
where $R_n^{(1)}$ is the relevance at the input layer calculated by Eqs.~\eqref{eq:define_rij_dtd_zplus} and~\eqref{eq:define_rij_dtd_zbeta} with the exception of the final layer relevance $R_{n}^{(L)}$ as:
\begin{align}
    R_{n}^{(L)}=
    \begin{cases}
        z_t^{(L)} & n=t  \\
        -\frac{z_t^{(L)}}{N-1} & \mathrm{otherwise}. 
    \end{cases}
    \label{def_dual_clrp}
\end{align}
In this way, the relevance of the target class becomes disentangled from the relevance of the other classes. 
This allows for the visualization of regions which contribute to the target class with the other classes penalized. 

Another property enforced by CLRP is that for the output layer $L$, the propagated relevance from the target class is equal to the sum of the penalty from the other classes, or:
\begin{align}
    R_{t}^{(L)} + \sum^{N-1}_{n | n \neq t} R_{n}^{(L)} = 0.
    \label{def_equal}
\end{align}
This balance between the target class and the other classes ensures that the relevance of the non-target classes do not overpower the relevance of the target class. 
This is especially useful for instances where there are many possible classes.

\begin{figure}[t]
        \centering
    \subfigure[Forward]{
        \centering
        \begin{minipage}{0.47\columnwidth}
        \centering
            \includegraphics[width=.9\columnwidth]{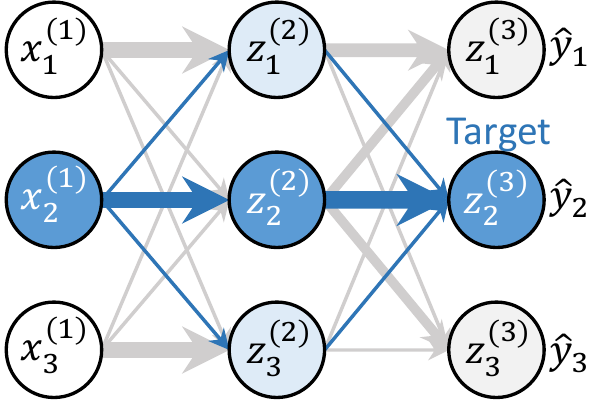}
    \vspace{1mm}
        \end{minipage}
    }
    \subfigure[Relevance Propagation]{
        \centering
        \begin{minipage}{0.47\columnwidth}
        \centering
            \includegraphics[width=.9\columnwidth]{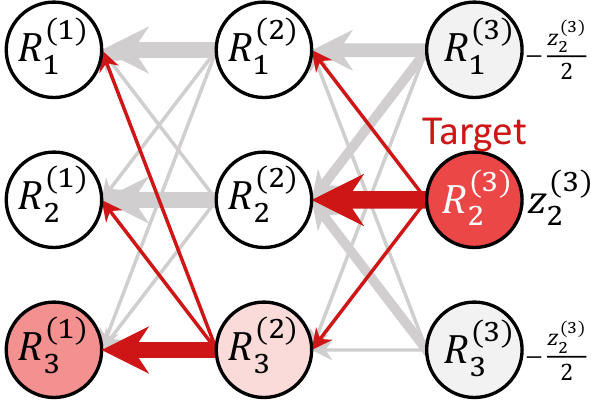}
    \vspace{1mm}
        \end{minipage}
    }
    \caption{In this example, the subtractive relevance of $R_1^{(3)}$ and $R_3^{(3)}$ incorrectly prevent the propagation to $R_2^{(2)}$. }
    \label{sglrp_merit}
\end{figure}

However, as shown in Fig.~\ref{sglrp_merit}, by equally penalizing non-target classes, problems may arise. 
In this example, $z_2^{(2)}$ has a strong correspondence with output $\hat{y}_2$. 
However, due to the penalties from the two non-target classes, the relevance $R_2^{(2)}$ is reduced to zero. 
Even worse, in the end, the relevance is miss-attributed to $x_3^{(1)}$ and not $x_2^{(1)}$ as it should. 
The cause of this phenomenon is due to the equal weighting of the non-target nodes.

\section{Softmax-Gradient Layer-wise Relevance Propagation}\label{proposal}

In order to achieve a more robust visualization of the pixel-wise class discriminate contribution that the input has on classification, we propose a novel method called SGLRP. 
LRP and DTD-based methods use the pre-softmax intermediary summation $z_t^{(L)}$ of output layer $L$ as the initial relevance $R_t^{(L)}$ of the target node $t$. 
Conversely, SGLRP is designed to take advantage of the post-softmax probabilities.  
As shown in Fig.~\ref{sglrp_target_figure}, 
specifically, we propose using the gradient of the softmax output $\hat{y}_t$ with respect to the intermediate value of each output node $z_n$ as the relevance of the output layer $R_n^{(L)}$. 
Namely, the relevance map $\bm{R}_{\mathrm{SGLRP}}$ is defined as: 
\begin{equation}
    \bm{R}_{\mathrm{SGLRP}} = \max\left(R_1^{(1)},\dots,R_n^{(1)},\dots, R_N^{(1)},0\right),
    \label{sglrp_def}
\end{equation}
where $R_1^{(1)},\dots,R_n^{(1)},\dots, R_N^{(1)}$ are the relevance values at the input layer calculated by Eqs.~\eqref{eq:define_rij_dtd_zplus} and~\eqref{eq:define_rij_dtd_zbeta}, except with the output layer relevance $R_{n}^{(L)}$ as the gradient of softmax, or:
\begin{align}
    R_{n}^{(L)} = \frac{\partial \hat{y}_t}{\partial z_n} =
    \begin{cases}
        \hat{y}_t(1-\hat{y}_t) & n=t \\
        - \hat{y}_t\hat{y}_n & \mathrm{otherwise}, 
    \end{cases}
    \label{partial_softmax}
\end{align}
where $\hat{y}_n$ is the post-softmax predicted probability of class $n$ and $\hat{y}_t$ is the probability of target class $t$. 
The derivation for the gradient of softmax can be found in the Supplementary Materials. 
By using the gradient of softmax as the initial relevance from the output layer, we create an LRP model which can propagate values that relate to directly to the probability that the object is that class. 
Furthermore, this is a more natural method of removing the relevance from the non-target classes compared to LRP which just ignores the other classes and CLRP which uses an arbitrarily fixed penalty. 

Another advantage of using the gradient of softmax maintains the property in Eq.~\eqref{def_equal} that the relevance of the target node $R_t^{(L)}$ is equal to the sum of all the other nodes. 
This means that SGLRP also has a balance between the target class and the other classes. 
However, unlike CLRP, due to using $- \hat{y}_t\hat{y}_n$ in Eq.~\eqref{partial_softmax} for the non-target classes, the interfering or adversarial objects will have a higher subtractive relevance meaning that they are specifically targeted for removal. 
For example, given an image with two objects of different classes, SGLRP will specifically remove the relevance from the non-target second class more heavily than classes that are not present in the image. 
This is opposed to LRP which does not subtract from any conflicting objects and CLRP which uses a uniform penalty. 


\section{Experimental Results} \label{evaluations}

In this section, the proposed method is compared to existing methods by qualitative and quantitative evaluations.

\subsection{Dataset and Architecture}\label{environments}
For the experiment, we used a Visual Geometry Group 16-layer CNN (VGG16)~\cite{Simonyan14c} trained using the ImageNet Large Scale Visual Recognition Challenge 2012 (ILSVRC2012)~\cite{Russakovsky2015} dataset. 
VGG16 contains two convolutional layers of 64 nodes, two convolutional layers of 128 nodes, three convolutional layers with 256 nodes, three convolutional layers with 512 nodes, an additional three convolutional layers with 512 nodes, two fully-connected layers with 4,096 nodes, and the softmax output layer. 
All of the convolutional layers use $3 \times 3$ stride 1 convolutions and the convolutional layer blocks are separated by $2 \times 2$ stride 2 max pooling. 
This network was used due to its generic but deep CNN structure and its high accuracy of $69.63\%$. 
It should be noted that the proposed method can be applied to any arbitrary neural network including other deep CNNs. 

The dataset used in the experiment was the ImageNet ILSVRC2012 dataset~\cite{Russakovsky2015}. 
It has 1,000 classes and about 1.2 million total images. 
The images are resized and cropped to $224 \times 224 \times 3$.
For the quantitative evaluation, the relevance map was prepared for 50,000 test set images.


\subsection{Qualitative Evaluation}\label{contrastive_visualization}

We first perform a qualitative evaluation based on the relevance maps produced by comparative pixel-wise methods. 
For the qualitative evaluation, we compare SGLRP to the related methods of LRP~\cite{Bach2015} and CLRP~\cite{gu2018understanding} as well as a state-of-the-art pixel-wise visualization method, Guided Grad-CAM~\cite{Selvaraju2017}. 
In the relevance maps, only the positive contributions are shown and they are normalized by $\frac{1}{\max |\bm{R}^{(1)}|}$ for visualization purposes.

\begin{figure}[t]
        \centering
        \includegraphics[width=1.0\columnwidth]{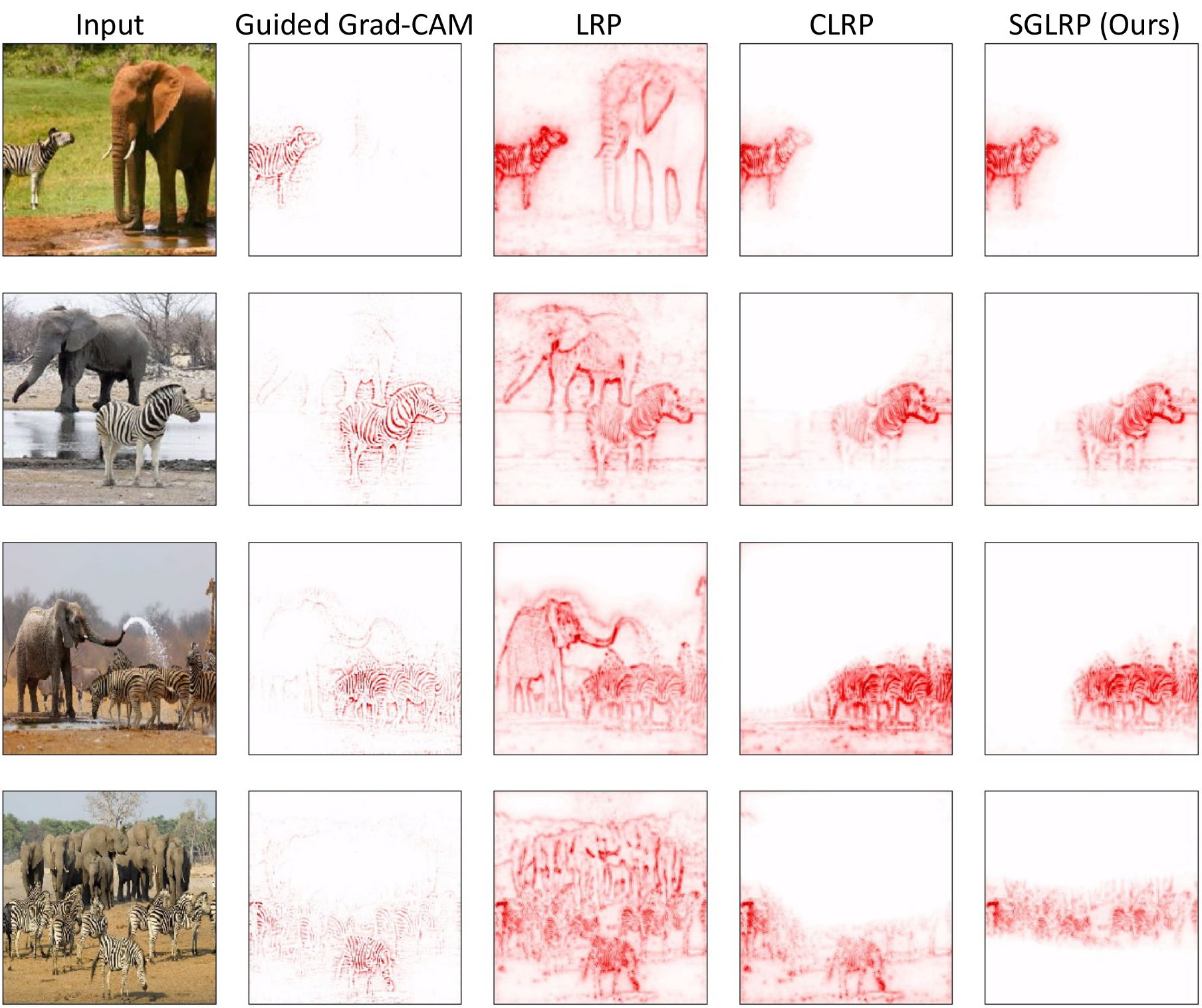}
    
    \caption{Visualizations of the relevance of ``Zebra."}
    \label{zebra}
\end{figure}



\subsubsection{Visualizations of Contrastive Classes}

A key feature of SGLRP is the ability to get the pixel-wise relevance maps for a target class. 
This target class does not necessarily have to be the predicted class. 
Therefore, we can compare the relevance maps of the four methods targeting any class which appears in the same image. 
Fig.~\ref{zebra} shows the relevance maps using ``Zebra" as the target. 
These images were chosen to match the qualitative evaluation used by Gu et al.~\cite{gu2018understanding}.

In the examples in Fig.~\ref{zebra}, LRP is able to determine the general salience of the object but is unable to specifically separate out the objects of the target class. 
In other words, not only is the relevance from the non-target class highlighted, many regions in the background are deemed to have relevance. 
Therefore, generally, LRP is insufficient for determining the relevance of objects.


Alternatively, CLRP and SGLRP are truly class discriminate and are able to correctly identify the relevance of the pixels which lead to the target object. 
Furthermore, by examining the fine details of Fig.~\ref{zebra}, we can confirm that the proposed method is able to isolate the relevance of the target objects with a higher accuracy than CLRP. 
Not only are the objects more defined, but there is less background noise and overlap between the object classes. 
This is possible because CLRP indiscriminately reduces the relevance of all classes, whereas SGLRP specifically removes objects found in the image that are not the target class.

\begin{figure}
        \centering
    \subfigure[Spider Web]{
        \includegraphics[width=1.0\columnwidth]{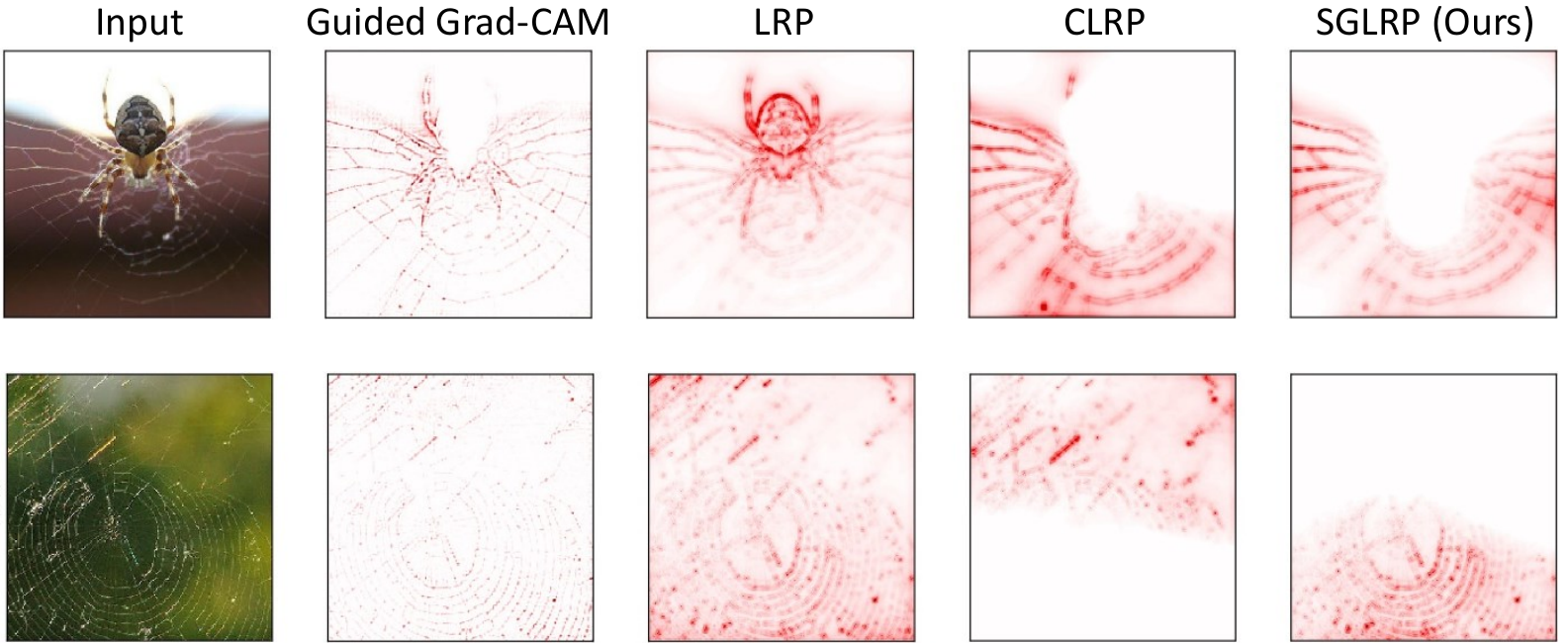}
    }
    
    \subfigure[Submarine]{
        \includegraphics[width=1.0\columnwidth]{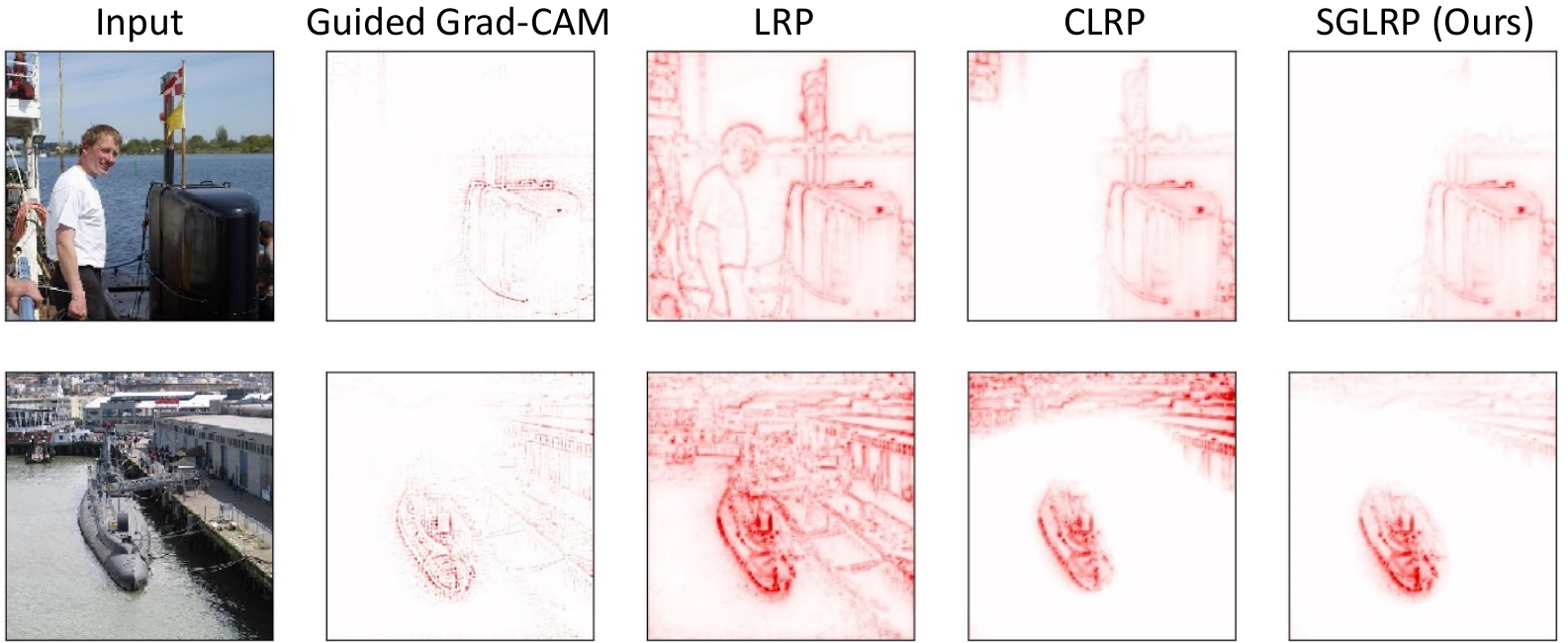}
    }
    
    \subfigure[Banjo]{
        \includegraphics[width=1.0\columnwidth]{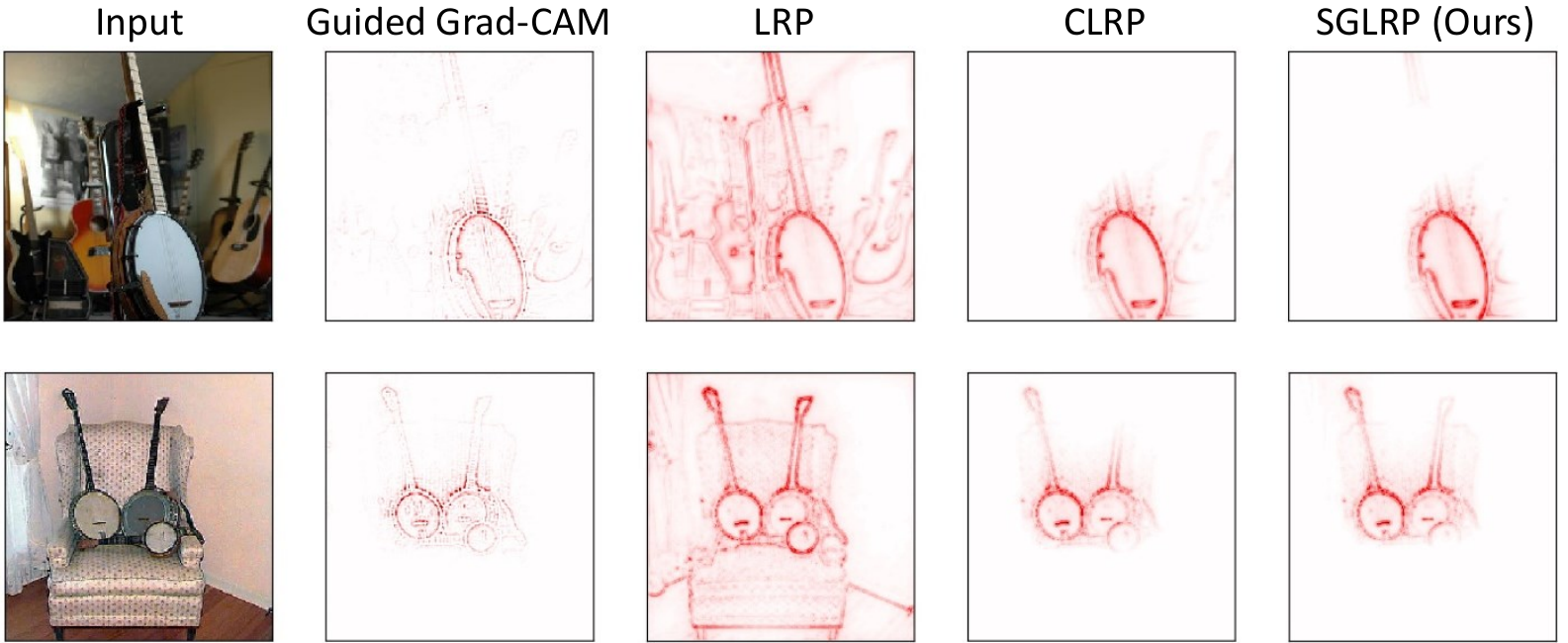}
    }
    
    \subfigure[Plastic Bag]{
        \includegraphics[width=1.0\columnwidth]{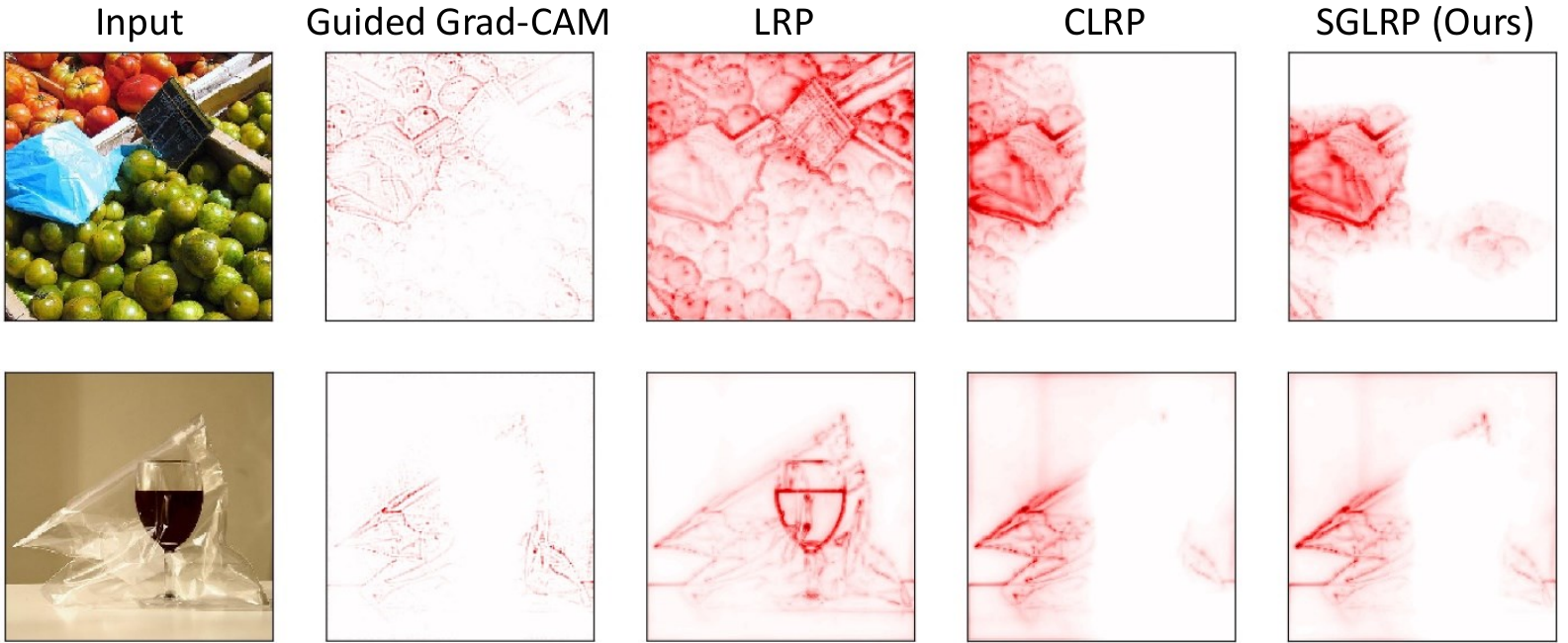}
    }
    
    \subfigure[Water Ouzel]{
        \includegraphics[width=1.0\columnwidth]{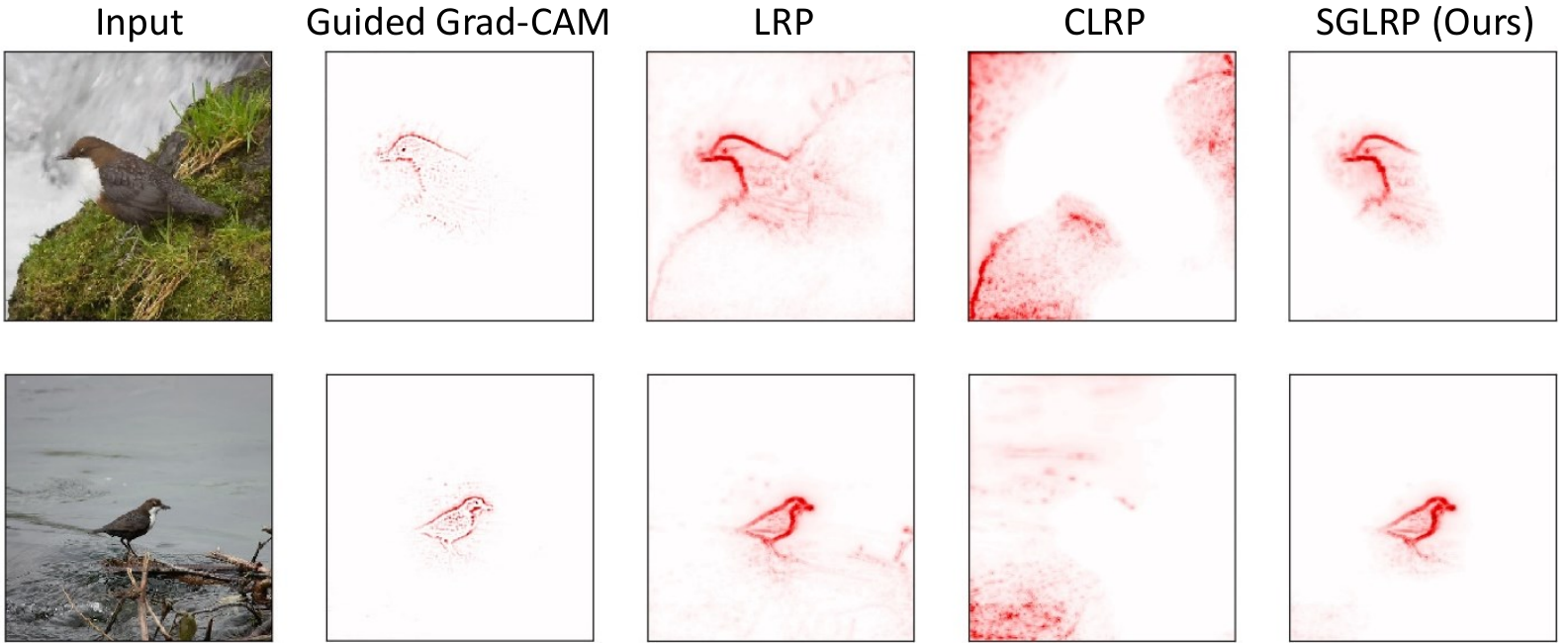}
    }
    \caption{Example relevance maps for various target classes. }
    \label{lrpexamples}
\end{figure}


\subsubsection{Visualizations Using Other Models}

\begin{figure}[t]
    \centering
    \includegraphics[width=1.0\columnwidth,trim={0cm 4.4cm 0cm 0cm},clip]{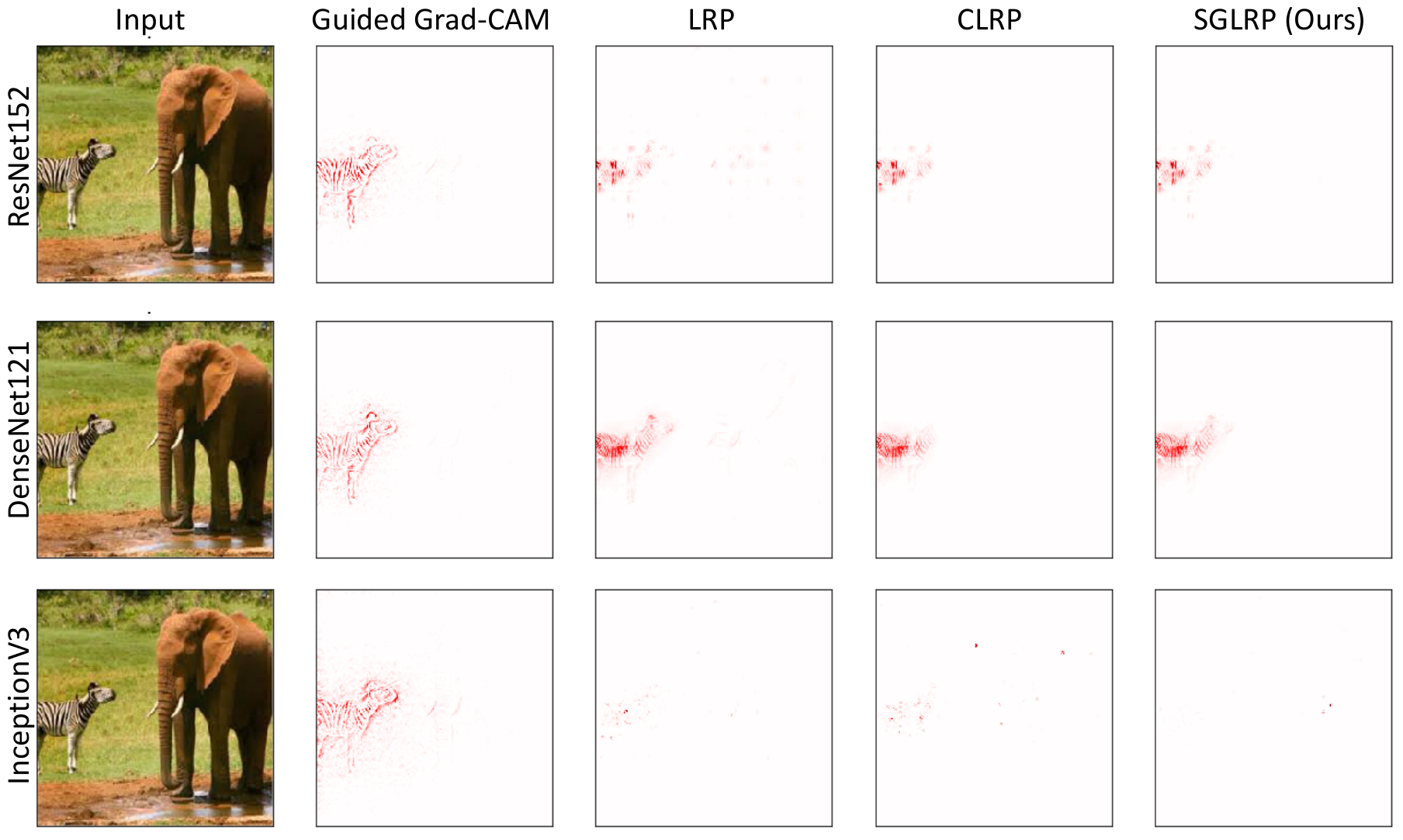}
    \caption{Comparisons on other networks targeting ``Zebra."}
    \label{othernetworks}
\end{figure}

The proposed method is a general visualization method and can be used with most neural network models. 
For example, Fig.~\ref{othernetworks} shows examples from ResNet152~\cite{He_2016}, DenseNet121~\cite{Huang_2017}, and InceptionV3~\cite{Szegedy_2016}.
In the examples, the results are similar to results using VGG16 in Fig.~\ref{zebra} with exception to the visualizations from InceptionV3. 
For all of the LRP-based methods, InceptionV3 has very sparse relevance maps. 
This may be attributed to the complex structure of InceptionV3.

\subsubsection{Performance Targeting the Ground Truth Class}

In this evaluation, we show examples of images from ILSVRC2012 using the ground truth label as the target. 
Fig.~\ref{lrpexamples} shows comparative examples for various classes in the dataset. 
The examples follow the same trend in that the classic LRP is insufficient for relevance visualization. 
At times LRP seems to only identify the general salience of the input, whereas Guided Grad-CAM, CLRP, and SGLRP provide a better picture of what is important for the classification of the VGG16 for the target class. 
The most interesting observation from the figure is how well SGLRP is at removing the conflicting class. 
For instance, in the ``Spider Web" example, the spider is almost entirely cut out from the relevance. 
The same thing happens for the person and buildings in the ``Submarine" class and the guitars and chair in the ``Banjo" class. 

Also, SGLRP tends to have an advantage over CLRP during the times where there are many classes similar to the target class in the dataset. 
For example, Fig.~\ref{lrpexamples} shows ``Water Ouzel"s but the ILSVRC2012 dataset has about 60 different classes of birds. 
CLRP seems to have more difficulty during these cases.

\begin{figure}
        \centering
    \subfigure[Odometer]{
        \includegraphics[width=1.0\columnwidth]{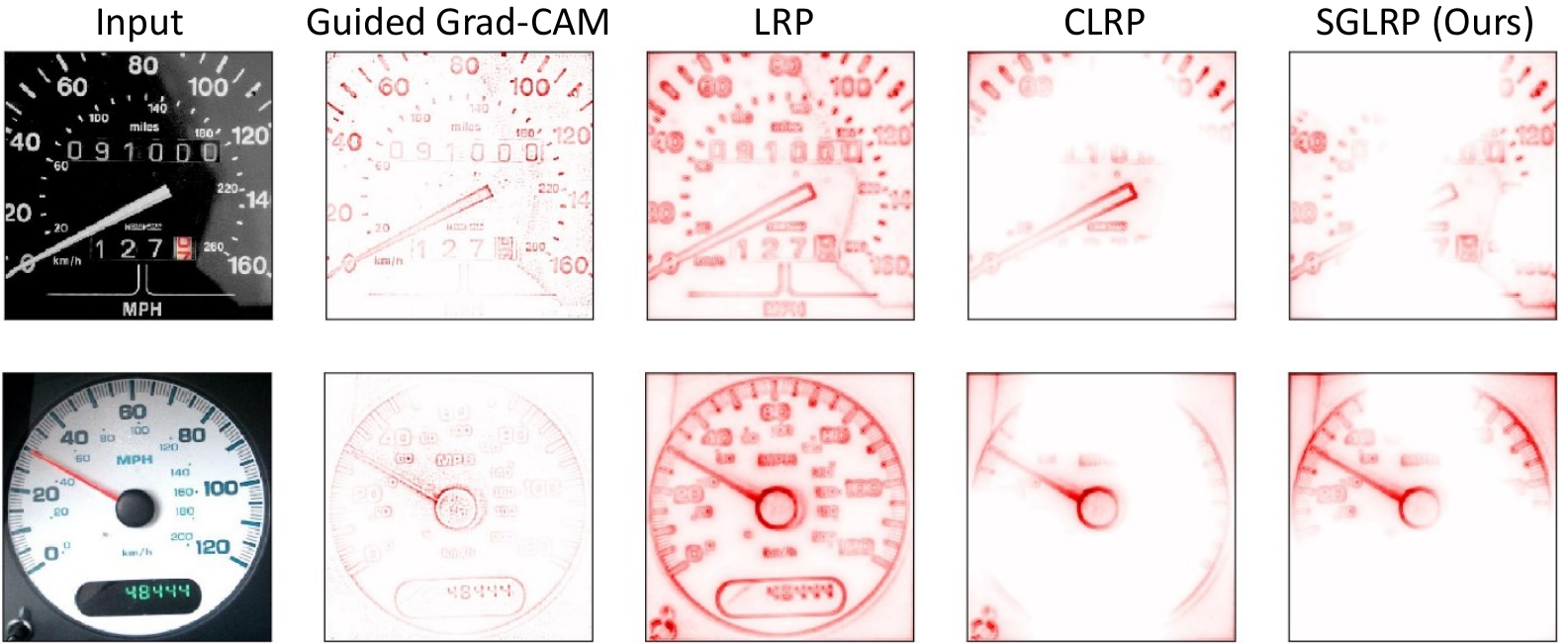}
        \vspace{-1.5mm}
    }
    
    
    \subfigure[Chocolate Sauce]{
        \includegraphics[width=1.0\columnwidth]{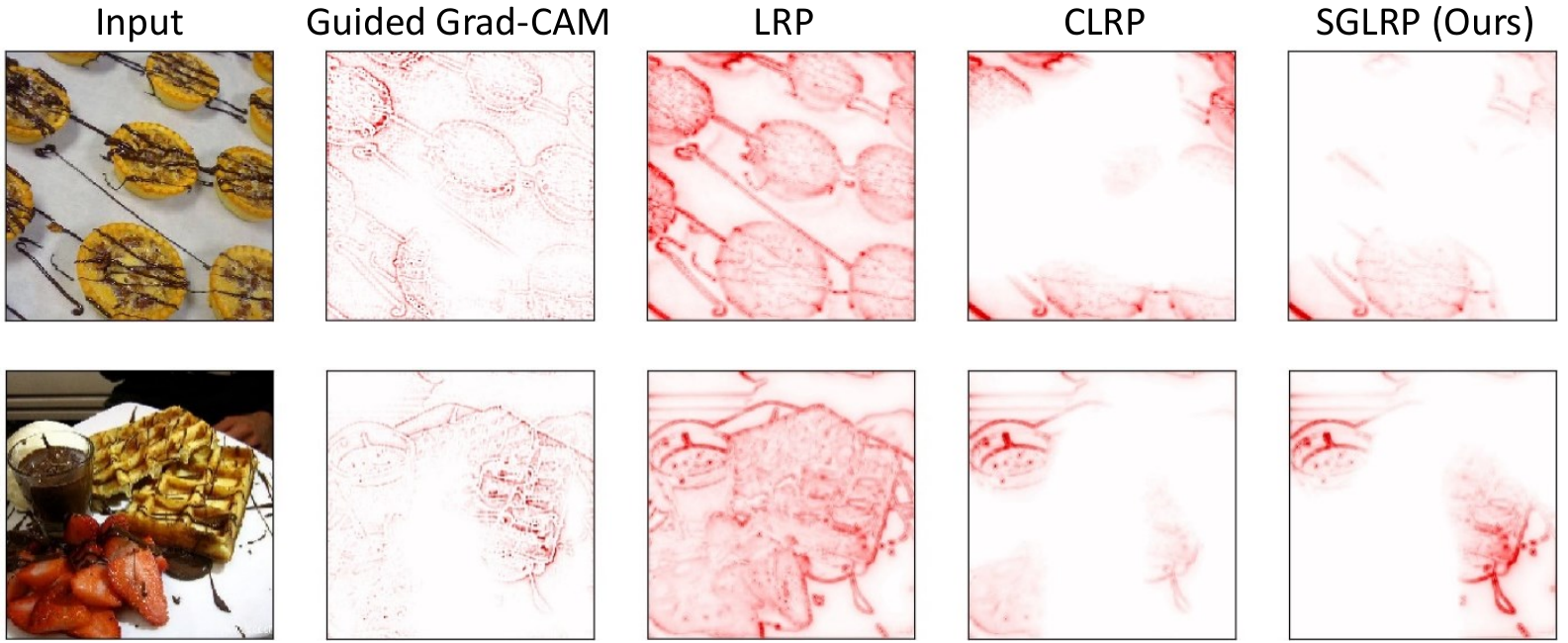}
        \vspace{-1.5mm}
    }
    \caption{Examples of bad results of SGLRP. }
    \label{badexamples}
\end{figure}

Fig.~\ref{badexamples} reveals instances where SGLRP performed poorly. 
SGLRP tends to do poorly when the target object encompasses the entire image, such as Fig.~\ref{badexamples}~(a). 
Another case where SGLRP has trouble is when the target object is not a significant part of the image, such as ``Chocolate Sauce" in (b). 
The reason for this failure is because the non-target occluding classes overwhelm the target class. 
Likewise, this problem is consistent with the results from CLRP. 

More examples are shown in the Supplementary Materials.

\subsection{Quantitative Evaluation}\label{ablation_study}

In order to quantitatively evaluate the proposed method, we performed three evaluations, maximal patch masking~\cite{gu2018understanding} of the ground truth class, maximal patch masking of the second most probable class, and the Pointing Game~\cite{Zhang_2016}. 
Four baselines are used for the quantitative evaluations, LRP~\cite{Bach2015}, CLRP~\cite{gu2018understanding}, Guided Grad-CAM~\cite{Selvaraju2017}, and Random. 
For the Pointing Game, Random is a relevance map of random points and for the maximal patch masking evaluations, Random is a random center point for the patches. Random is used as a baseline to demonstrate the strength of deliberate approaches.

\subsubsection{Maximal Patch Masking of the Ground Truth Class}
\label{sec:evalgt}

\begin{figure}[t]
    \centering
    \includegraphics[width=1.0\columnwidth]{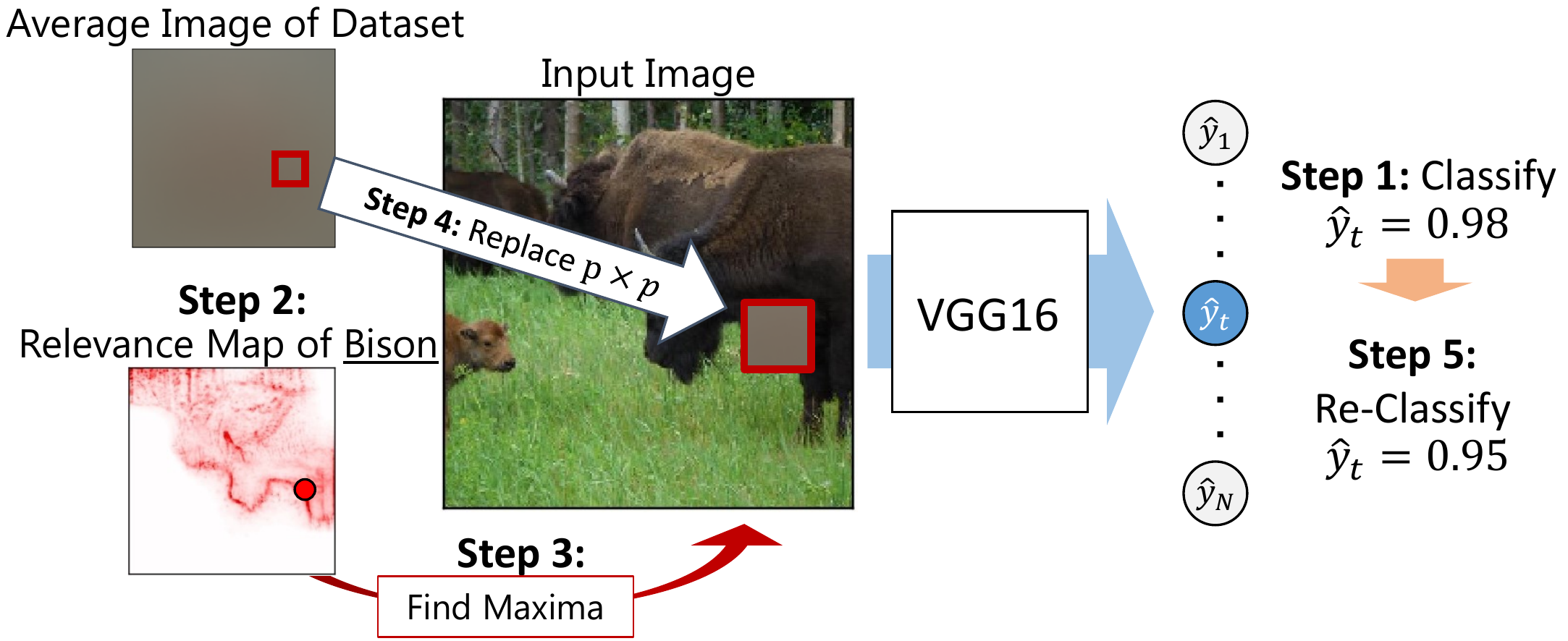}
    \caption{Demonstration of the maximal patch masking evaluation. After determining the relevance map, the maximal value and surrounding pixels are masked and the classification is repeated. Through this, the change in the output probability is observed.}
    \label{ablation_study_figure}
\end{figure}

In this evaluation, we measure the amount that the target node's output $\hat{y}_t$ drops when the maximally relevant patch is removed from the input image, as shown in Fig.~\ref{ablation_study_figure}. 
This evaluation was proposed by~\cite{gu2018understanding}. 
The general idea is that regions considered important to the prediction of the target class should have a high relevance. 
Therefore, by removing the patch surrounding the highest relevance, the probability of the target class should drop. 
Under this scheme, larger $\hat{y}_t$ drops would mean that the removed patch was more important for the classification, thus more relevant. 
Thus, by measuring the change in $\hat{y}_t$, it is possible to quantitatively evaluate the different visualization methods.


The maximal patch masking evaluation is performed in five steps. First, the image is classified using VGG16 and the value of the predicted outcome $\hat{y}_t$ of the ground truth class is recorded. 
Second, relevance maps from LRP, CLRP, SGLRP, and Guided Grad-CAM is calculated. 
Third, the maximal relevance value on the relevance map with respect to the ground truth class is found. 
Fourth, a patch of size $p \times p$ surrounding the maximal value on the input image is replaced by a patch of the average values of the dataset. 
The extra baseline of Random is used by selecting a random point on the input image and masking a patch as if it were a maximal value. 
Finally, the image with the masked region is re-classified and the drop in $\hat{y}_t$ is monitored.

\begin{figure}[t]
    \centering
    \subfigure[Ground Truth as the Target]{
    \includegraphics[width=0.47\columnwidth]{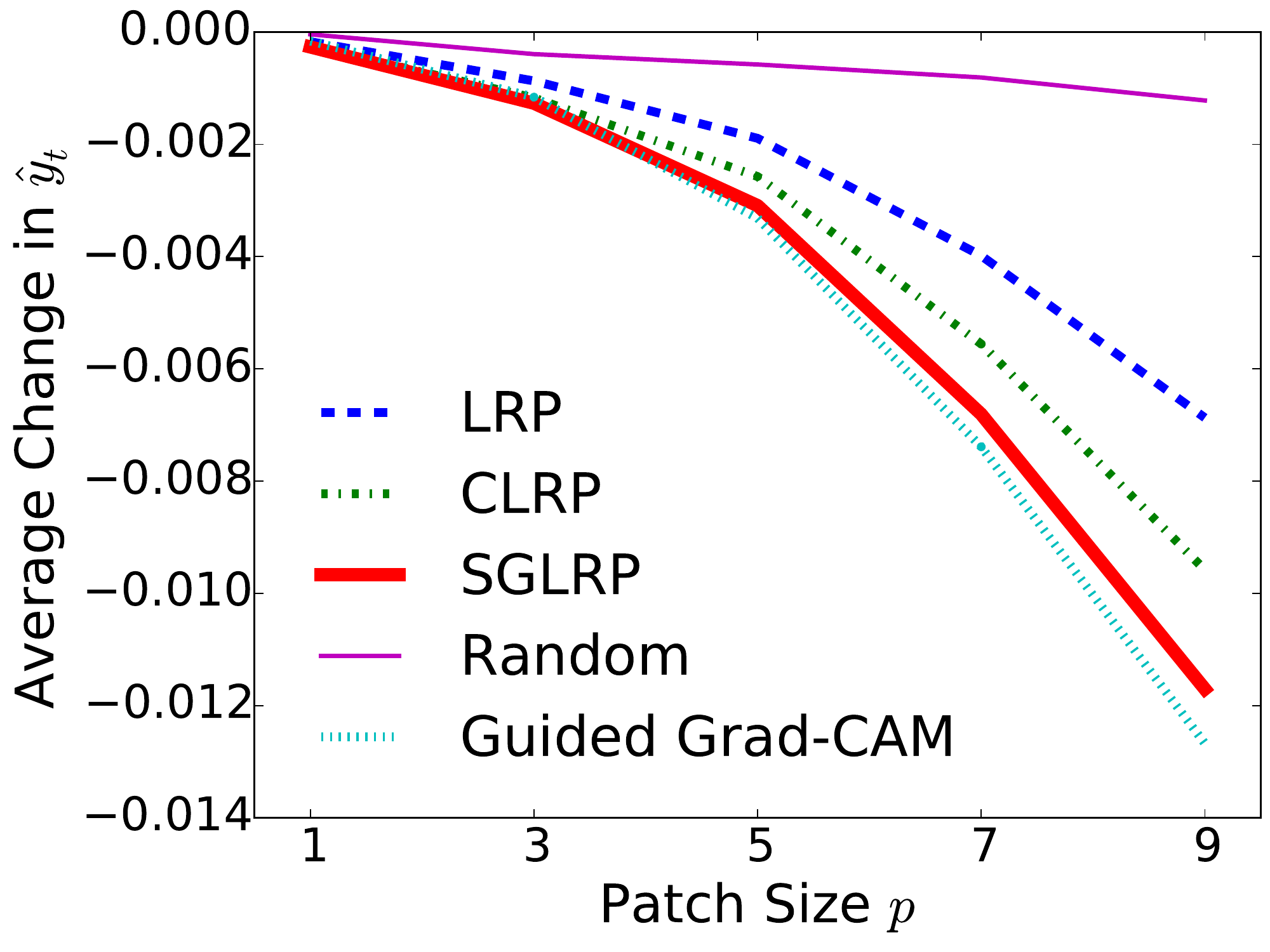}
    }
    \subfigure[2nd Probable Class as the Target]{
    \includegraphics[width=0.47\columnwidth]{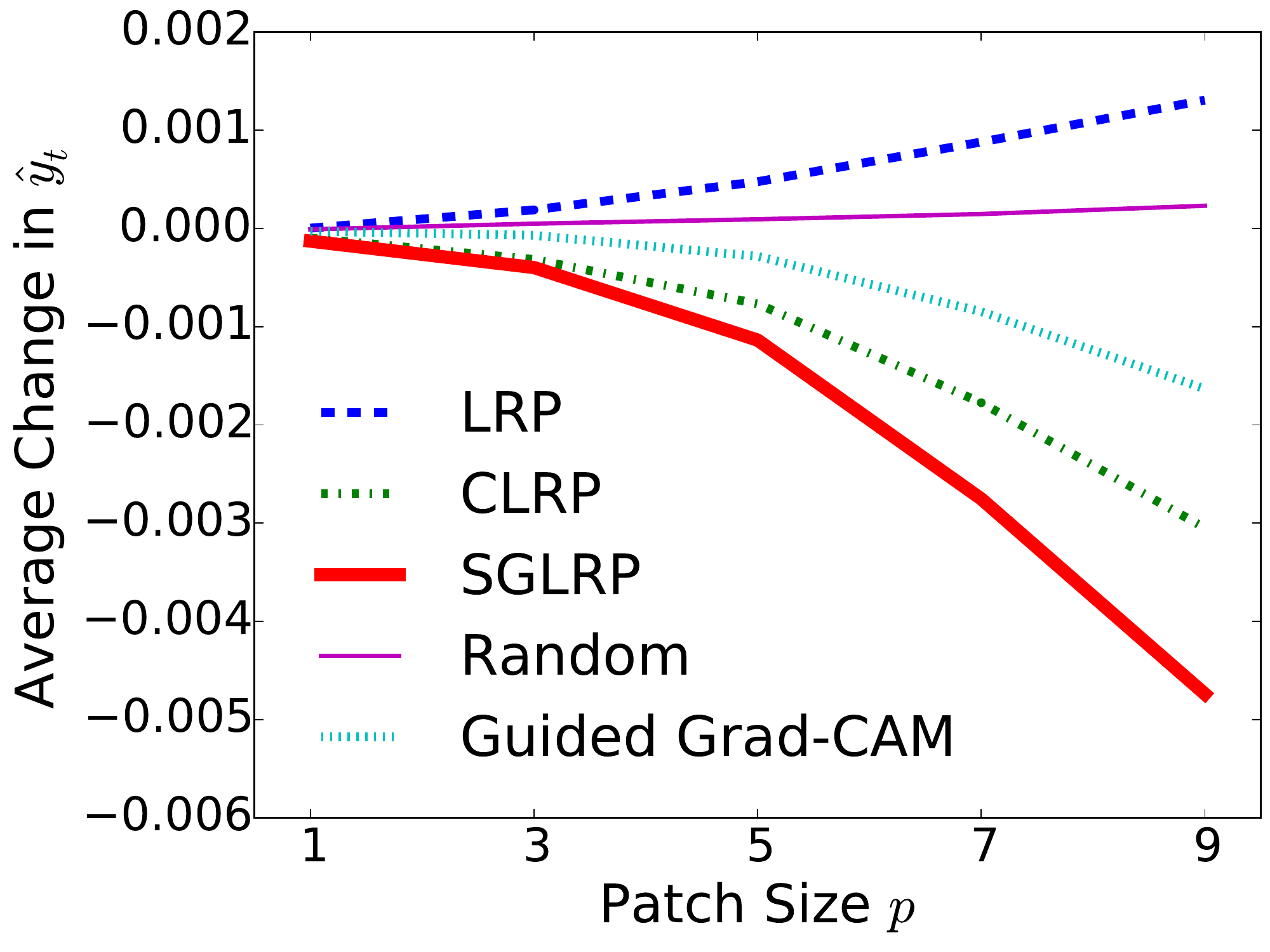}
    }
    \caption{Change in the value of $\hat{y}_t$ when masking the (a) ground truth class and the (b) class with the second highest probability. Lower is better. }
    \label{ablation_gt_yt}
\end{figure}

The results in Fig.~\ref{ablation_gt_yt} (a) show that the effects of removing the most relevant region determined by each method with different patch sizes, $p=1,3,5,7,9$. 
The results show that maximal patch masking reduced the value of $\hat{y}_t$ for SGLRP more than any other LRP-based method across every patch size. 
This confirms that SGLRP outperforms LRP and CLRP in identifying the regions most relevant for the target class's classification. 
However, Guided Grad-CAM the value of $\hat{y}_t$ dropped slightly more for $p \geq 5$. 

\subsubsection{Maximal Patch Masking of the Second Probable Class}
 
Given the ILSVRC2012 dataset, the most salient object has a high probability of being the ground truth object. 
Consequently, there is a possibility that monitoring the drop in prediction using the ground truth as the target only tests the method's ability to find the most salient object. 
However, SGLRP's main advantage is the ability to be class discriminative.

Therefore, to assess SGLRP's ability to be class discriminative, we performed a second evaluation targeting the class with the second highest probability. 
Using the second highest probability allows for the evaluation of target objects that are not the primary label of the image, for example in Fig.~\ref{lrpexamples}, the spider in (a), the person in (b), the guitars in (c), etc. 
Another reason for using the second highest probability and not the first is because the VGG16 model used in the experiments has a $69.63\%$ with Top 1 accuracy. 
This means that for $69.63\%$ of the test images, the first highest is the same as the ground truth, which we explored in Section~\ref{sec:evalgt}. 

Fig.~\ref{ablation_gt_yt} (b) shows the results when using the same conditions as Section~\ref{sec:evalgt}, except with the second highest probable class instead of the ground truth label. 
Unlike using the ground truth, this evaluation showed that SGLRP had the best performance across all patch sizes. 
The results also show that Guided Grad-CAM performed relatively poorly and was even below CLRP across all patch sizes. 
This indicates that while SGLRP and CLRP can discriminate target classes, Guided Grad-CAM is more limited to only the dominant, most salient class. 
It is also interesting to note that removing the maximally relevant regions increased the likelihood of the target class for LRP and Random, which implies that the maximal (or random) point lies in the primary class.

\subsubsection{Pointing Game}

\begin{figure}[t]
    \centering
    \includegraphics[width=1.0\columnwidth]{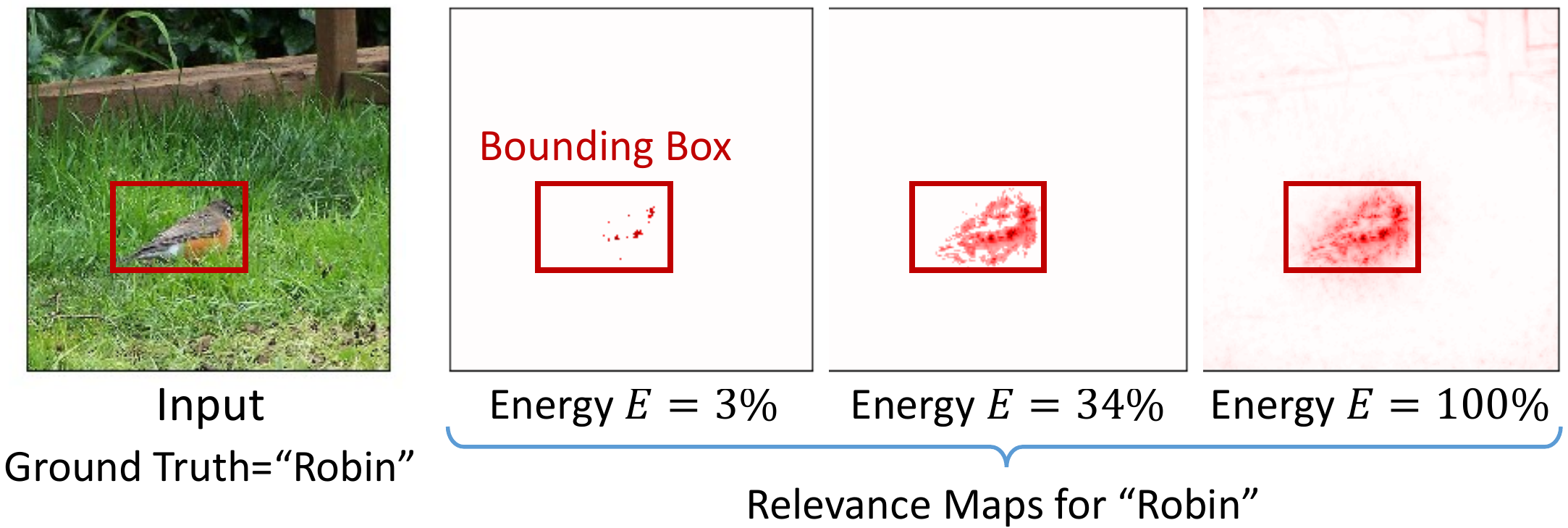}
    \caption{Illustration of the Pointing Game. Provided a bounding box, the Pointing Accuracy is determined by the ratio of pixels inside the bounding box at different energy levels.}
    \label{pointing_game_def}
\end{figure}

The Pointing Game was proposed by Zhang et al.~\cite{Zhang_2016}, which tests the localization ability of visualization methods. 
Given an object bounding box provided by the dataset, the Pointing Game determines the accuracy at which the maximal point lies within the bounding box. 
Gu et al.~\cite{gu2018understanding} found that the original Pointing Game was too forgiving, thus they extended the game to include all relevant points. 
We use this the extended Pointing Game, as shown in Fig.~\ref{pointing_game_def}. 
Specifically, the Pointing Accuracy is calculated by whether a point on the relevance map lies within the bounding box of the target object, or:
\begin{equation}
    Pointing Accuracy = \frac{\#Hits}{\#Hits+\#Misses}.
    \label{eq:acc}
\end{equation}
A $Hit$ is counted if a relevance value above threshold $\tau$ falls within the provided bounding box and a $Miss$ would be a relevance value above $\tau$ which lies outside the bounding box. 
The threshold $\tau$ is determined by the percent energy $E$ in the relevance map. 
$E$ is defined as:
\begin{align}
    E = \frac{\# \left( R_t^{(1)} \geq \tau \right)}{\#  \left( R_t^{(1)} > 0 \right)}.
    \label{eq:energy}
\end{align}
Note, Energy $E=100\%$ would be a threshold of $\tau=0$ and would include every positive value on the relevance map which would a very low Pointing Accuracy.

\begin{figure}[t]
    \centering
    \includegraphics[width=1.0\columnwidth]{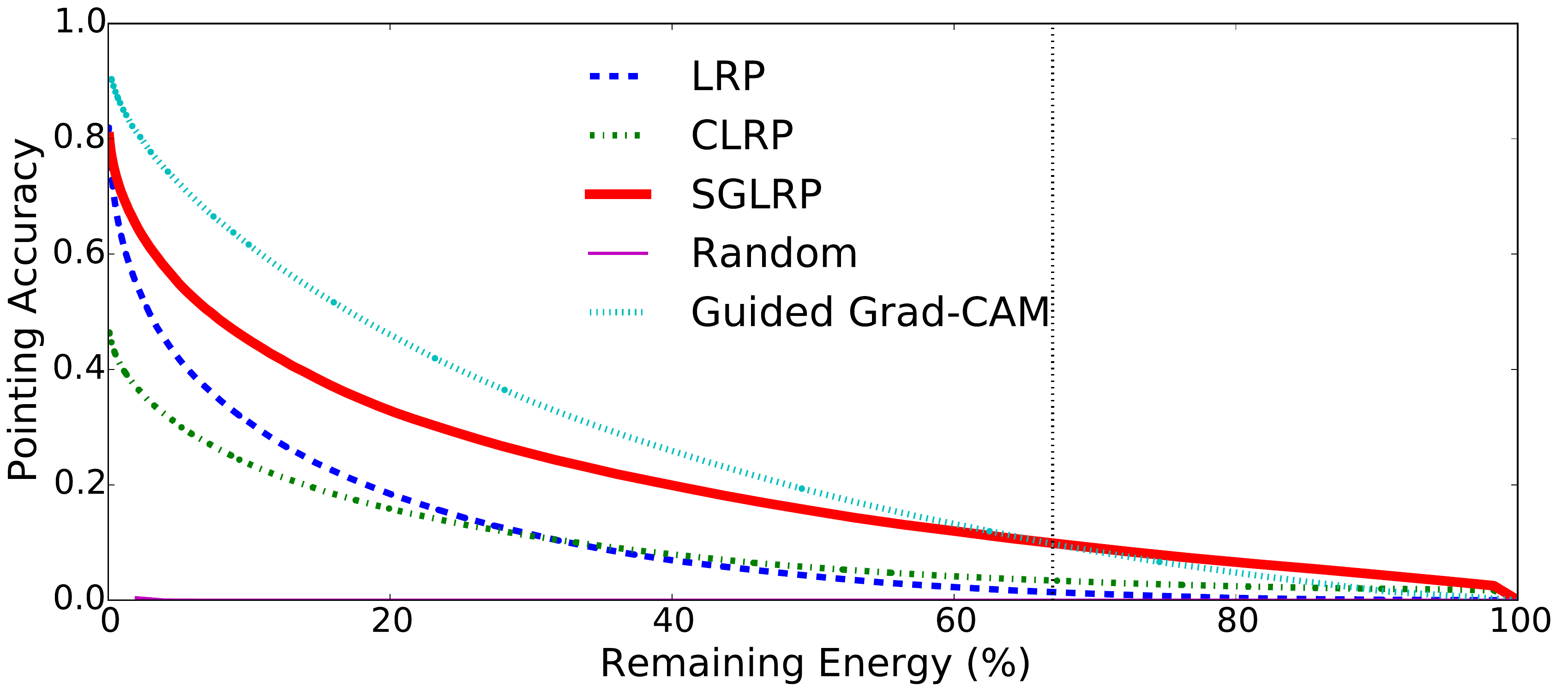}
    \caption{Pointing Accuracy of the comparative models at different thresholds determined by amount of energy. Higher is better. The vertical line indicates the point at which SGLRP overtakes Guided Grad-CAM.}
    \label{pointing_accuracy}
\end{figure}

The results of the Pointing Game is shown in Fig.~\ref{pointing_accuracy}. 
In the results, the Random baseline had a near-zero Pointing Accuracy for all energy levels. 
As for SGLRP, it had the highest Pointing Accuracy out of all of the LRP-based methods. 
On the other hand, Guided Grad-CAM had a higher accuracy for $E<67\%$. Consequently, for the energy levels $E\geq 67\%$, SGLRP had the highest Pointing Accuracy. 
A high Pointing Accuracy at high energy ratio means that there were less identified relevant pixels outside of the bounding box. 

\section{Conclusion}

In this paper, we proposed a novel method for understanding the contribution that regions in the input have on classification predictions. 
The proposed Softmax-Gradient Layer-wise Relevance Propagation (SGLRP) is an extension of LRP which adds the ability for class discrimination by subtracting the relevance from non-target classes using the gradient of softmax.  
In order to compare the effectiveness of the proposed method with existing methods, we performed qualitative and quantitative evaluations. 
Through qualitative and quantitative analysis, we demonstrated the strength that SGLRP has at attributing input regions that lead to object classification. 

In the future, this research can be a useful tool for understanding the underlying mechanisms in a neural network's design process. 
It can also be extended to any neural network structure and be used for many applications such as localization and visualization. 
An implementation of SGLRP using the iNNvestigate~\cite{alber2019innvestigate} library can be found at
\url{https://github.com/uchidalab/softmaxgradient-lrp}.

\clearpage

{\small
\bibliographystyle{ieee}
\bibliography{contents/ref}
}
\end{document}